\pdfoutput=1
\documentclass[11pt]{article}

\usepackage[preprint]{main}

\usepackage{amsmath,amsfonts,bm}

\def\eqref#1{equation~\ref{#1}}

\def\1{\bm{1}}

\DeclareMathAlphabet{\mathsfit}{\encodingdefault}{\sfdefault}{m}{sl}
\SetMathAlphabet{\mathsfit}{bold}{\encodingdefault}{\sfdefault}{bx}{n}

\usepackage{times}
\usepackage{latexsym}
\usepackage[T1]{fontenc}
\usepackage[utf8]{inputenc}
\usepackage{microtype}
\usepackage{inconsolata}
\usepackage{graphicx}

\title{Tracking Universal Features Through Fine-Tuning and Model Merging}

\author{Niels Horn \\
  Department of Computer Science \\
  University of Copenhagen \\
  \texttt{niels@horn.ninja} \\\And
  Desmond Elliott \\
  Department of Computer Science \\
  University of Copenhagen \\
  \texttt{de@di.ku.dk} \\}

\usepackage{xspace}
\newcommand{\bp}{\texttt{BabyPython}\xspace}
\newcommand{\lua}{\texttt{Lua}\xspace}
\newcommand{\ts}{\texttt{TinyStories}\xspace}
\newcommand{\ls}{\texttt{LuaStories}\xspace}

\begin{document}
\maketitle
\begin{abstract}
We study how features emerge, disappear, and persist across models fine-tuned on different domains of text. More specifically, we start from a base one-layer Transformer language model that is trained on a combination of the BabyLM corpus \cite{2301.11796}, and a collection of Python code from The Stack \cite{2211.15533}. This base model is adapted to two new domains of text: TinyStories \cite{tiny_stories}, and the Lua programming language, respectively; and then these two models are merged using these two models using spherical linear interpolation. Our exploration aims to provide deeper insights into the stability and transformation of features across typical transfer-learning scenarios using small-scale models and sparse auto-encoders.

\end{abstract}

\section{Introduction}

Language models are proving useful on an ever-widening range of tasks but there are still open questions about what is actually learned and represented in these models. Researchers have invested substantial energy into understanding language models, from both behavioural~\cite{ribeiro-etal-2020-beyond} and mechanistic perspectives~\cite{elhage2021mathematical}. More recently, \citet{bricken2023monosemanticity} proposed to study neural network features as a form of dictionary learning, in which a feature is a function that assigns values to data points. Their features are extracted from a sparse autoencoder using the activations of the MLP layer in a Transformer block. This approach to understanding what is learned by language models has proven successful, and can even control the output of language models pinning the features~\cite{templeton2024scaling}.


In this paper, we study language model feature \emph{evolution}, i.e. the emergence, disappearance, and persistence of features. We study feature evolution in two common transfer-learning settings: fine-tuning a language model to a new domain~\cite{gururangan-etal-2020-dont}, and merging the weights of two models~\cite{weight_averaging}. We conduct our experiments using a small language model trained from scratch on a combination of English text and the Python programming \mbox{language}.\footnote{We believe that programming languages offer an interesting test-bed for feature evolution because on the one hand, they contain reserved keywords that overlap with common English words, e.g. ``try'', ``while'', ``break'', etc; while on the other hand they have sequences not commonly seen in English text, such as code indentation blocks, e.g. ``\texttt{\textbackslash t}'', or ``\hspace{4ex}''. We study Python and Lua in this paper but we expect that our methods can be extended to other programming languages.} This model is then separately fine-tuned on a different programming language (Lua), and more English text, after which the the fine-tuned models are merged back into a single model. Given this family of related models, we use sparse auto-encoders to extract and correlate feature activation patterns to study the evolution of features. We find that very few features persist between the studied models, but those that do persist are interpretable, e.g. they correspond to generic properties of the text, such as punctuation and formatting. We report case studies on a persistent feature that represents variable assignments in programming languages, and a disappearing feature that handles exceptions.

\section{Related Work}
\label{related}

Feature universality and convergence in Transformer language models is a growing area of interest. Prior work shows that models converge at universal feature representations~\cite{convergent_learning,universality, convergent_learning, platonic_hypothesis}, and more recent work shows these features can be extracted across different and divergent models \cite{bricken2023monosemanticity, templeton2024scaling, sparse_autoencoders_find_interpretable_features}. In \citet{bricken2023monosemanticity}, feature universality is the ability of a sparse auto-encoders to extract universal features across similar and divergent models. 

Common approaches to adapting language models include fine-tuning on new data~\cite{gururangan-etal-2020-dont}, and model merging \cite{weight_averaging, model_fusing, model_soups}, especially for practical use cases \cite{mergekit}. However, little is known about the features in fine-tuned and merged models compared to their starting point. Our work extends \citet{bricken2023monosemanticity} to study the feature dynamics of model merging, and fine-tuning, and further explore feature universality across models.

\section{Methodology}

The goal of our study is to investigate the evolution of Transformer language model features in common transfer-learning settings. In contrast to recent work on frontier models~\cite{templeton2024scaling,gao2024scaling}, we use one-layer Transformers, which are quick to train and reproducible in lower-compute environments. Inspired by \citet{bricken2023monosemanticity}, our models are Mistral-like \cite{mistralReport} Transformers, with 1 self-attention block, followed by a 1024D MLP with a SiLU  activation~\citep{elfwing2018sigmoid}. 
The language model features are extracted using a sparse autoencoder trained on the MLP activations. Each language model and sparse auto-encoder can be trained in 10 hours, and 8 hours, respectively, on an NVIDIA A100 40GB GPU.\footnote{Our code and weights will be available upon publication.} 

\begin{figure}[t]
\centering
\includegraphics[width=0.49\textwidth]{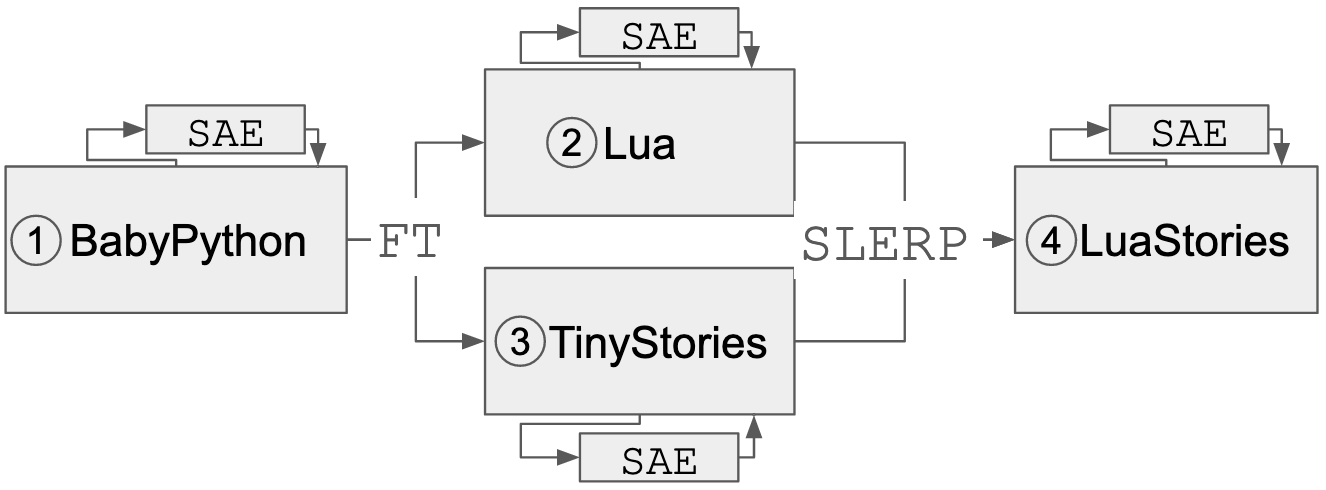}
\vspace{-1em}
\caption{Overview of the experimental design. We start with a base model trained on BabyLM and Python code (1), which is fine-tuned (FT) on two new domains: the Lua programming language (2), and TinyStories (3). The fine-tuned models are merged into a single LuaStories model using spherical linear interpolation (SLERP) interpolation (4). For each of these models, we train a sparse auto-encoder on the MLP activations using the same data distribution as the original model.}\label{fig:exp_design}
\end{figure}

\subsection{Model Training and Fine-Tuning}\label{sec:model_training}

The base model, \bp, is trained on a dataset containing the BabyLM 100M corpus \cite{2301.11796} and a 10\% sample of the Python subset of The Stack \cite{2211.15533}, resulting in even amounts of Python and BabyLM tokens.\footnote{335.6M tokens in total, according to the model tokenizer.}
We fine-tune two further models: one on the Lua subset of The Stack, and the other on the TinyStories dataset \cite{tiny_stories}. These fine-tuned models, named \lua and \ts, respectively, are then merged into a fourth model, \ls. All models are trained with an autoregressive log-likelihood loss. Figure~\ref{fig:exp_design} shows an overview of the relationship between the models.

\subsection{Autoencoder Feature Extraction}
\label{sec:sparse_autoencoder}

The learned features used in our study are extracted from a sparse autoencoder, which is trained with an expansion factor of 16 on the output MLP activations of each Transformer language model.\footnote{We follow best-practices, including tracking ``dead'' neurons in the autoencoder using the L0 norm, finding no significant problems. See Appendix \ref{app:sparse_autoencoders} for details.} We follow the general approach of \citet{bricken2023monosemanticity}, using 15 million tokens of 24-token blocks sampled uniformly from the combined trace of datasets used to train each respective model.\footnote{For example, the \ls sparse auto-encoder is trained on data sampled from the combined BabyPython, Lua, and TinyStories datasets. Whereas the \lua sparse auto-encoder is trained on data sampled from the combined BabyPython and Lua datasets.} This process allows us to extract features from all of the underlying training distributions of each model. The sparse autoencoders are used to gather feature activation patterns, which are used to correlate individual extracted features across models, allowing us to observe the evolution of the features.

\subsection{Model Merging}

Given that all models start from the same base parameters, we can use generalised linear mode connectivity \cite{linear_mode_connectivity} to enable model merging via spherical linear interpolation. Model merging techniques are used to combine multiple pre-trained models into a unified model that retains what is learned by the original models.
We use spherical linear interpolation\footnote{We experimented with other merging methods, such as TIES-merging \cite{TIES}, but none worked as well as Slerp. We leave further study of this for future work.} \cite{Shoemake1985AnimatingRW}, to interpolate all model parameters along a spherical path between the parameters of the Lua model to the parameters of the TinyStories model. At every fraction $t$ along the path we measure the accuracy of the model corresponding to the interpolated parameters on both the Lua and TinyStories validation data, as measured by correct next-token prediction. This allows us to select the merged model that retains the optimal balance between each model. Figure~\ref{fig:merged_model_accuracy} in Appendix~\ref{app:slerp_curves} shows the accuracy of the merged model at each fraction $t$ along the interpolated path is shown. We pick the model corresponding to the parameters at equilibrium of the merged model accuracy on Lua and TinyStories at $t=58\%$. This LuaStories model's accuracy is 20\% lower than both original models in their respective domain, but approximately 20\% better than the shared base model of the two original models. The LuaStories model is equally accurate in modelling Lua and TinyStories data.

\subsection{Quantifying Feature Evolution}\label{sec:measuring_evolution}

We quantify the evolution of features extracted from the sparse autoencoders of pairs of language models: a \emph{parent} model and a \emph{child} model. Following  \citet{bricken2023monosemanticity}, features are \emph{similar} if they take similar values over a diverse set of data, which is calculated by collecting feature activation patterns from the sparse autoencoder of each model.\footnote{We collect activations over 3 million tokens of data sampled from the common data distribution (see Footnote 5).} We define a feature as \emph{persisting} if there exists a feature pair with activation patterns that correlate more than 80\% between the parent and child models.\footnote{The correlation threshold of 80\% is based on several rounds of manual inspection. We find that sparse features with activation patterns that correlate more than 80\% across are qualitatively similar enough to be considered the same.} A feature is \emph{emerging} when we cannot find any feature in the parent model that correlates more than 80\% with the features in the child model. And a feature is \emph{disappearing} if there are no features in the child model that correlate sufficiently with a feature in the parent model. 

\section{Empirical Results}

\subsection{High-level Feature Flow}
\label{feature_lineage}

\begin{figure}[t]
    \begin{center}
    \includegraphics[width=1.0\linewidth]{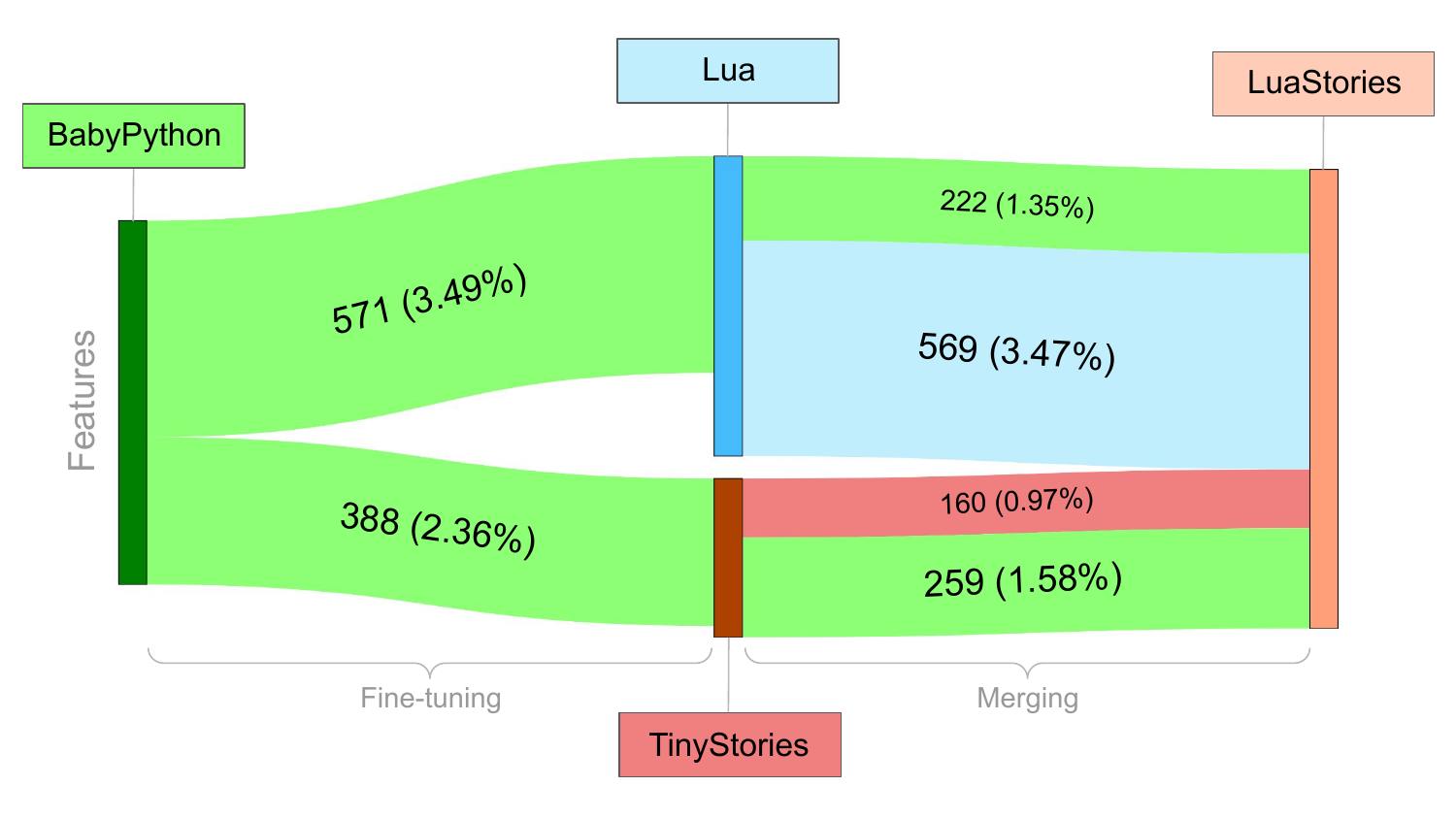}
    \end{center}
    \caption{Overview of the features persisting through fine-tuning and model merging, showing volumes and trajectories of extracted features that emerge, persist and disappear. This overview omits the features that don't persist, and so the visual flows are scaled proportional to the persisting features.  We note the share of features that persist from each model.}
    \label{fig:feature_flow}
\end{figure}

Figure~\ref{fig:feature_flow} shows an overview of the emerging, disappearing, and persisting features across the models. First, we note that most of the features from the base model \bp disappear through fine-tuning, i.e. only 959 features persist into either of the fine-tuned models. In the final merged model, \ls, there are 1,210 features that can be traced back to either the \lua or \ts models, of which 729 emerged during fine-tuning, while the remaining 481 features persisted from the initial \bp model.

We use automated interpretability to explore the features that do persist through to the final LuaStories model  (see Appendix~\ref{app:automated_interpretability}). By manually grouping verified explanations, we observe that 20\% of features persisting from \bp $\rightarrow$ \lua $\rightarrow$ \ls track punctuation, and 5\% can be explained as formatting features tracking indentation and different types of line breaks, which is a common property of the Python language. Although some features are seemingly monosemantically linguistic features, the remaining set of the features are less easily categorised. Overall, the majority of features flowing from the \bp model all the way to the \ls model are code related.

\subsection{Feature Flow Case Studies}

We now present a detailed examination of how two learned features evolve through our models. We study a feature that persists through both fine-tuning and merging, and we study a feature that disappears through fine-tuning. Feature behaviour is quantified using a proxy by checking the corresponding log-likelihood of a string under the under the hypothesised feature explanation, and under the full empirical activation distribution (see Appendix~\ref{app:llr_proxy} for more details).

\subsubsection{Persisting Variable Assignment Feature}
\label{assignment_feature}

\begin{figure}[t]
    \begin{center}
    \includegraphics[width=1.0\linewidth]{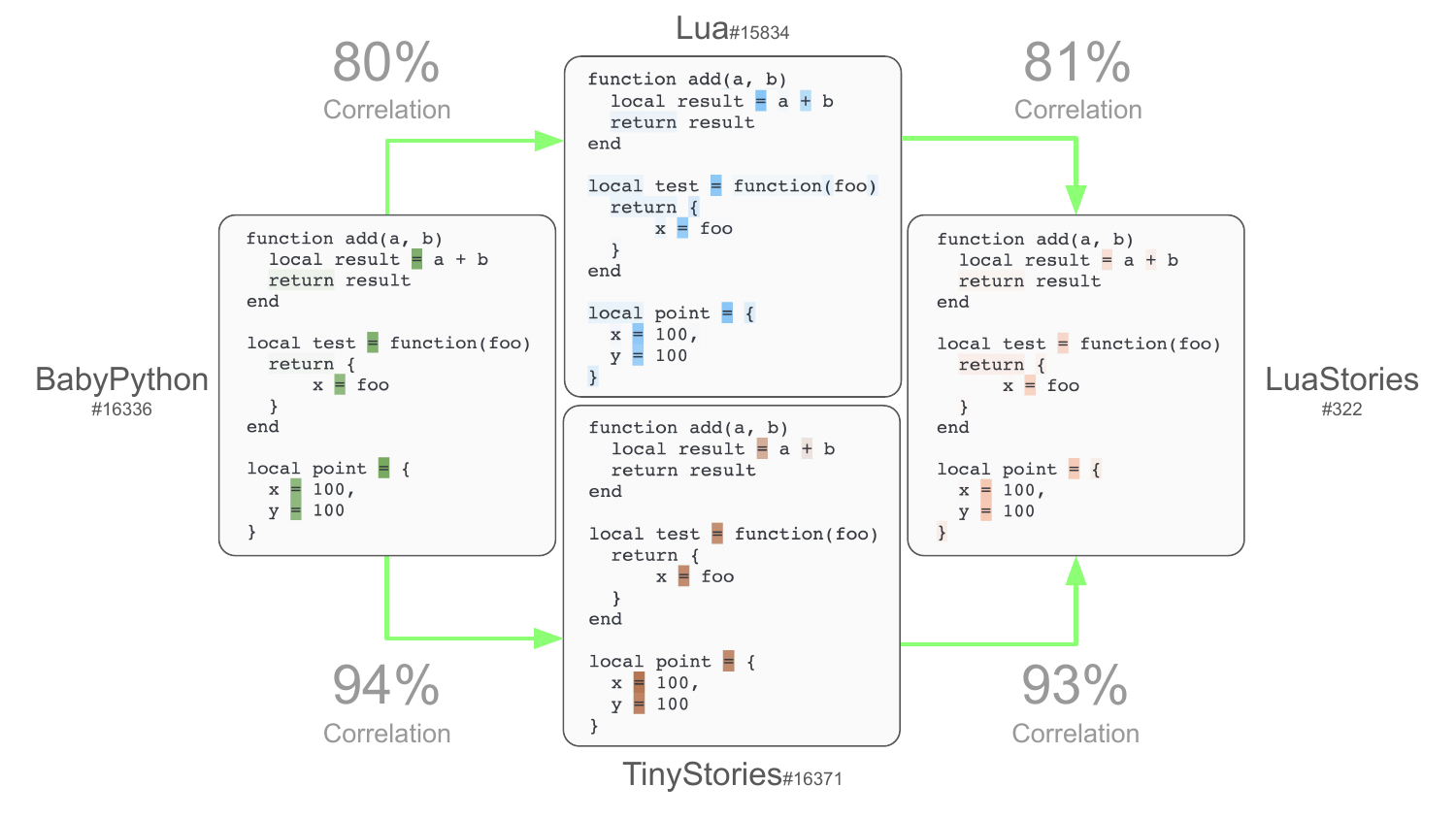}
    \end{center}
    \caption{Visualisation of the feature activation patterns of the universally extracted variable assignment features found in each model. Each token is highlighted according to the feature's activation level, where darker background colour denotes higher level of activation. Additionally, we note the observed activation pattern correlations between each feature.}
    \label{fig:assignment_feature}
\end{figure}

A clear and persisting feature across all models is the variable assignment feature, originally found in the \bp feature \#16336. This feature is activated by different types of variable assignment, as shown on a \lua example in Figure~\ref{fig:assignment_feature}. We sweep for corresponding assignment features across each subsequent model, to find the strongest correlating feature activation patterns in the \lua and \ts model, and in turn in the \ls model. By analysing all similar features, we find a feature in each of the models that seems to be close to identical to the assignment feature found in the \bp model.

We find that our tracked set of features are qualitatively universal, by analysing top-activating tokens and contexts, and through manual inspection. To further confirm this, we look at the log-likelihood ratio under the "\verb|=|" token feature hypothesis, and under the full empirical distribution computed as a 3 million token sample of the combined Lua and BabyPython dataset. Using this feature proxy, we find that the \bp feature has a log-likelihood ratio for assignments of 7.53, \lua has 6.58, \ts has 7.64, and \ls has 7.19. These features have a mean cross-correlation of 85.1\%. Therefore, we conclude that the assignment feature persists, as per our 80\% correlation threshold.

We observe the same  universality across all models, confirming that our sparse auto-encoders are able to extract the same universal feature across all models. This further reinforces the claims of extraction universality of \citet{bricken2023monosemanticity}.

\subsubsection{Disappearing Python Exception Feature}

As we fine-tune the base model to instead specialise solely on Lua code, many syntactical Python constructs become redundant. Intuitively, we expect this to be reflected in the extracted features. In the previous section, we are able to trace and recover an assignment feature from the \ts model. In this case we show the feature that tracks Python code exceptions disappears during fine-tuning.

The Python exception feature in the BabyPython model tracks components of  \verb|except| clauses in Python code. Using the activation data for the  Python exception feature in the BabyPython model, we sweep for similar activation patterns in the Lua model. Here we find a small handful of features.

\begin{figure}[t]
    \begin{center}
    \includegraphics[width=1.0\linewidth]{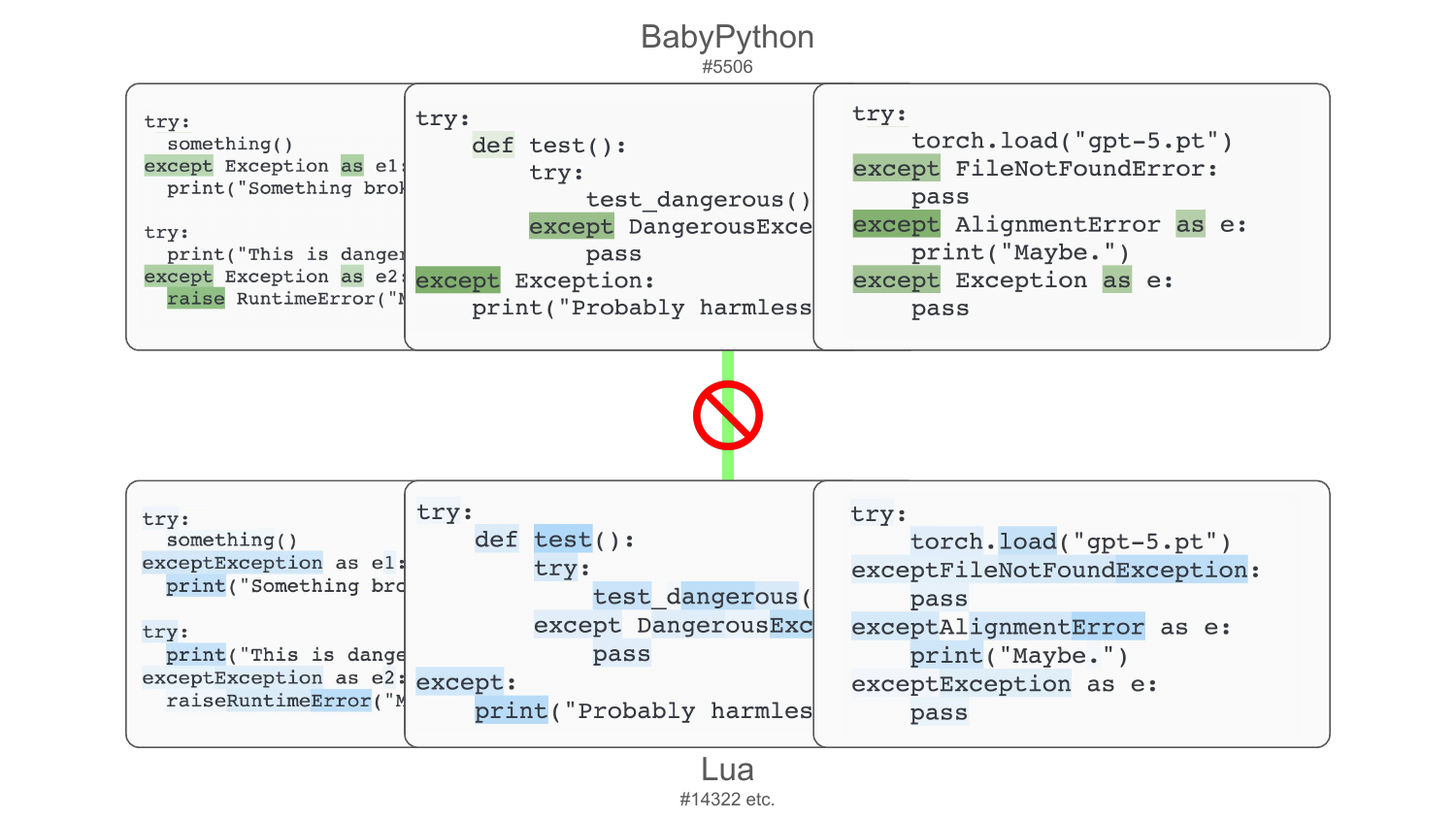}
    
    \end{center}
    \caption{Examples of observed activation patterns of the \bp Python exception feature, and the closest matching feature in the \lua model, qualitatively showing insufficient correlation between the two.}
    \label{fig:baby_python_except}
\end{figure}

Similar to §\ref{assignment_feature} we look at the log-likelihood ratios. Here, BabyPython \#5506 has a log-likelihood ratio of 11.38 under the simplified "\verb|except|" token hypothesis, and under the full empirical distribution. Lua \#14322 has a log-likelihood ratio of 8.59, and is 7 times more active for other tokens. These two features, while sharing some excitement for exceptions, correlate 40.81\% across observed activation patterns. As such, we find that this Python exception feature disappears through Lua fine-tuning, because we don't see any matches for the BabyPython~\#5506 feature.

\section{Conclusion}

Using sparse auto-encoders, we have empirically mapped the evolution of learned features in small Transformer language models through realistic transfer-learning scenarios. We find that spherical linear interpolation of model parameters is able to maintain features of the parent models. We show that features are diluted through model merging and fine-tuning and, in our experiments, don't correlate as well as they would for two similar Transformers trained with the same hyper-parameters on identical data. Future work includes scaling this to deeper language models, and to further exploration of feature evolution in other domains.

\section*{Limitations}

We conduct our experiments on just under half a billion tokens limited to the specific domains of English and programming languages, which may not capture feature evolution dynamics of larger and more diverse natural language corpora such as the Pile \cite{pile}.

\bibliography{main}

\appendix

\section{Slerp Interpolation curves}
\label{app:slerp_curves}

\begin{figure}[h!]
    \begin{center}
    \includegraphics[width=0.75\linewidth, trim={1cm 0 0 2cm},clip]{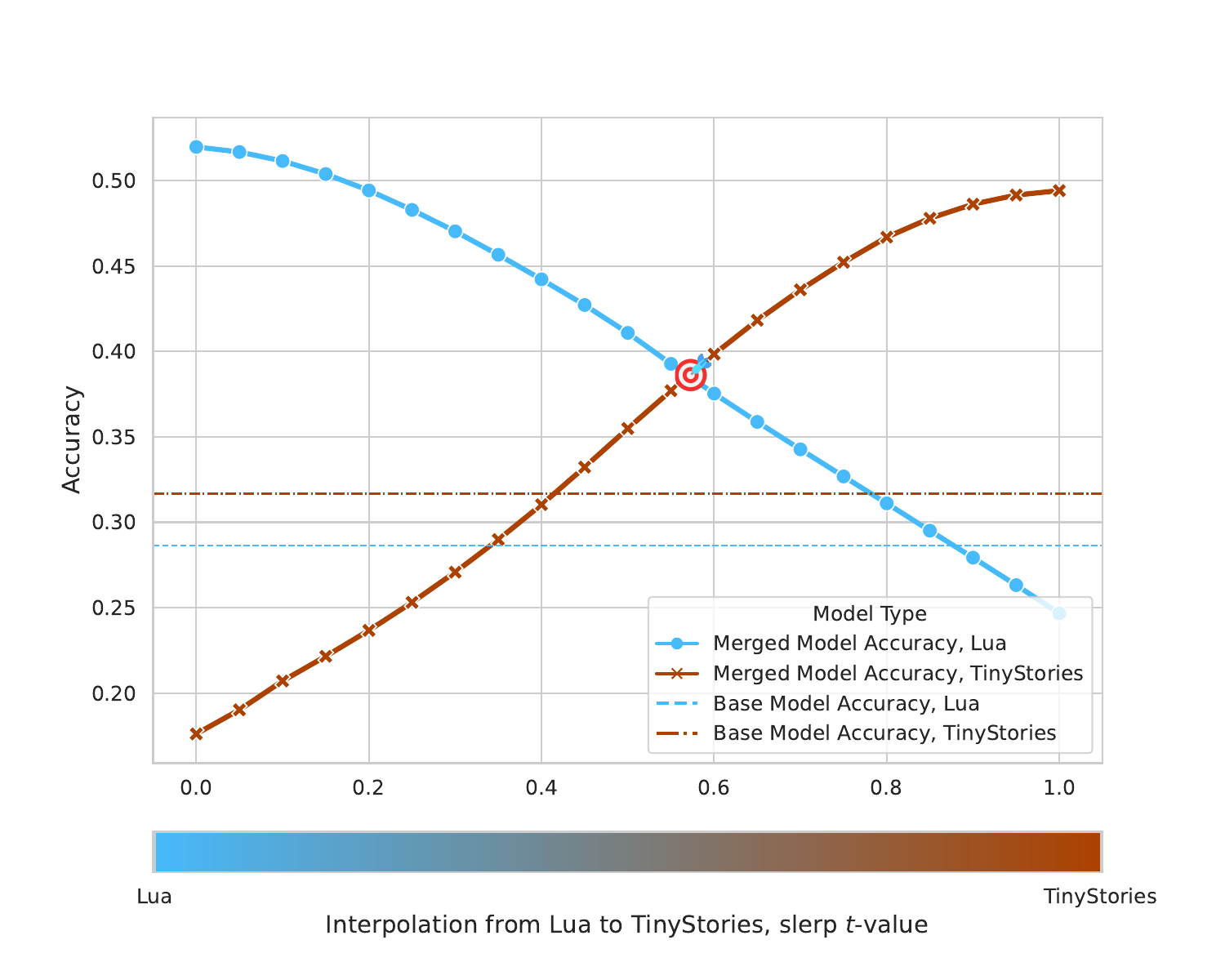}
    \end{center}
    \caption{Observed accuracy trends of a merged model consisting of weights spherically linear interpolated between the Lua model and the TinyStories model, as measured on the validation datasets of the Lua and TinyStories, respectively, at each interpolation step (slerp $t$-value). The dashed baselines show the accuracy of the shared base model underlying the Lua and TinyStories model, on the same validation datasets.}
    \label{fig:merged_model_accuracy}
\end{figure}

Figure~\ref{fig:merged_model_accuracy} shows the accuracy of different stages of the interpolation between the Lua and TinyStories models.

\section{Sparse Auto-Encoders and Universality}
\label{app:sparse_autoencoders}

We train our sparse-autoencoders according to directions in \citet{bricken2023monosemanticity}, using the Adam optimizer \cite{KingBa15} with $\beta_1 = 0.9$ and $\beta_2 = 0.9999$, a constant learning rate of $1e-4$, an L1-coefficient of $3e-4$, on activations corresponding shuffled blocks of 24 tokens in batches of 128 blocks. To extract activations, we use TransformerLens \cite{nanda2022transformerlens}, specifically targeting the \verb|blocks.0.mlp.hook_post| activations of the one-layer Mistral-like models. Our final sparse auto-encoders all have a mean L0 norm between 25-35 and mean MSE loss lower than 0.0067, explained loss of more than 88\% across all sparse auto-encoders, and were finally evaluated by manual and automated interpretability/explainability. For automated interpretability details see \ref{app:automated_interpretability}.

Formally, similar to \citet{bricken2023monosemanticity}, let $n$ be the input and output dimension of our sparse auto-encoder, and $m$ be the auto-encoder hidden layer dimension. Given encoder weights $W_e \in \mathbb{R}^{m \times n}$, $W_d \in \mathbb{R}^{n \times m}$ with columns of unit norm, encoder biases $\textbf{b}_e \in \mathbb{R}^m$, decoder biases $\textbf{b}_d \in \mathbb{R}^n$, the operations and the loss function over dataset $X$ are:

\begin{align}
    \bar{\mathbf{x}} &= \mathbf{x} - \frac{1}{|\textbf{x}|} \sum_{i=1}^N \mathbf{x}_i - \mathbf{b}_d \\
    \mathbf{f} &= \text{ReLU}(W_e \bar{\mathbf{x}} + \mathbf{b}_e) \\
    \hat{\mathbf{x}} &= W_d \mathbf{f} + \mathbf{b}_d \\
    \mathcal{L} &= \frac{1}{|X|} \sum_{\mathbf{x} \in X} \|\mathbf{x} - \hat{\mathbf{x}}\|_2^2 + \lambda \|\mathbf{f}\|_1
\end{align}

Our hidden layer is overcomplete, specifically $n=1024$ and $m=16n$, and the mean squared error is a mean over each vector element while the L1 penalty is a sum ($\lambda$ is the L1 coefficient). The hidden layer activations $\textbf{f}$ are the learned features of the sparse auto-encoder.

\section{Automated Interpretability}
\label{app:automated_interpretability}

To generate, evaluate and explore our extracted features at scale, we use automated interpretability using GPT-4 Turbo. We do this by sampling activation data similar to \citet{bricken2023monosemanticity} tasking the large language model to provide a concise explanation of the observed feature activation pattern, without including verbatim token nor or activation samples, based on a prompt containing sampled pairs of token observations and the corresponding quantised feature activation. This prompt format follows that of \citet{bills2023language}. To evaluate these explanations, we simulate the feature activation patterns using GPT-4 Turbo, on another similarly sampled set of feature activation pairs, and then measure the correlation between the simulated activations and the true observed activations. This is again similar to \citet{bricken2023monosemanticity} and \citet{bills2023language}. Using this approach to evaluate common features for each of our models, we arrive at Pearson correlation distributions that closely match those of \citet{bricken2023monosemanticity}.

To generate interpretations and evaluate a single feature costs approximately USD 0.11, using \texttt{gpt-4-turbo} via the OpenAI API. Therefore, it cost USD 185.68 to attempt to automatically explain all features in Figure~\ref{fig:feature_flow}.

\begin{figure}[h]
    \begin{center}
    \includegraphics[width=0.85\linewidth]{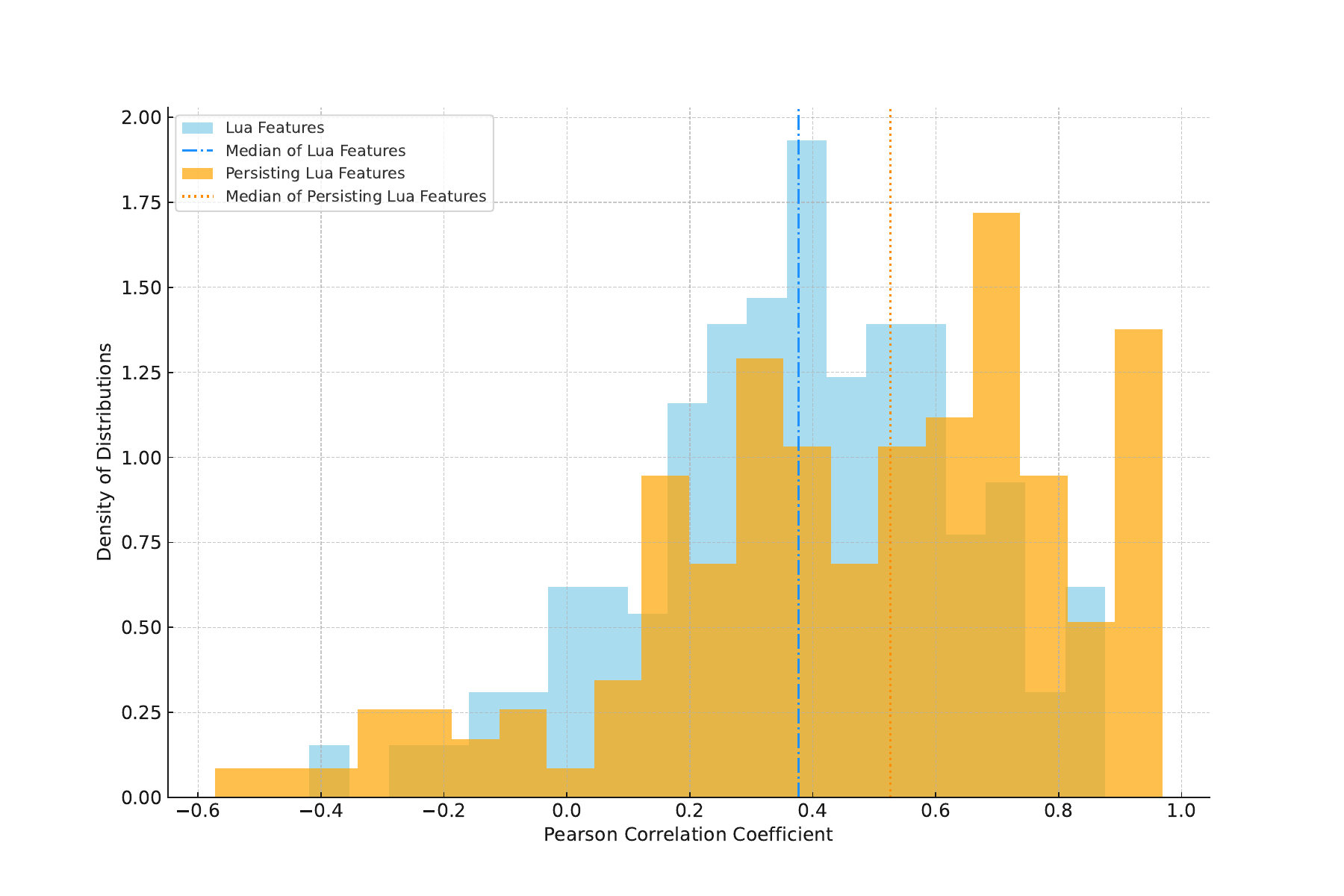}
    \end{center}
    \caption{Histograms showing the distribution of observed correlations between automatically generated explanations and the true feature activation patterns of features in the Lua and LuaStories model.}
    \label{fig:lua_explanations}
\end{figure}

In Figure~\ref{fig:lua_explanations} we show the correlation distributions used to evaluate the quality of our explanations of features in the Lua model. Here we distinguish between the features that persist through model merging to the LuaStories model and those that do not.

\section{Log-likelihood Ratio Feature Proxy}
\label{app:llr_proxy}

To analyse features, we adopt the log-likelihood ratio feature proxy used in \citet{bricken2023monosemanticity}. This is the log-likelihood ratio of a string under the feature hypothesis and under the full empirical distribution, where the feature hypothesis refers to a measure corresponding to the hypothesised explanation of a feature's activation pattern. For example, $\log\left(P\left(s|\text{variable assignment}\right) / P\left(s\right)\right)$. Following \citet{bricken2023monosemanticity}, we use log-likelihood proxies on the intuition that features will be incentivized to track log-likelihoods, since they linearly interact with the logits.

To compute $P(s)$, we use the unigram distribution over tokens and compute $P(s) = \prod_{t \in s} p(t)$. For the single-token features we deal with in the case studies of this work, we follow this same approach.

\end{document}